\documentclass[11pt]{article}

\usepackage{acl}
\usepackage{times}
\usepackage{latexsym}
\usepackage[T1]{fontenc}
\usepackage[utf8]{inputenc}
\usepackage{microtype}
\usepackage{inconsolata}
\usepackage{graphicx}
\usepackage{booktabs}
\usepackage{multirow}
\usepackage{amsmath,amssymb,amsthm}
\usepackage{enumitem}
\usepackage{xspace}
\usepackage{algorithm}
\usepackage{algpseudocode}
\usepackage{tabularx}
\usepackage{array}
\usepackage{url}

\newcommand{\method}{PCFJudge\xspace}
\newcommand{\pairmethod}{APOCJudge\xspace}
\newtheorem{proposition}{Proposition}

\title{Permutation-Consensus Listwise Judging for Robust Factuality Evaluation}

\author{
Tianyi Huang\thanks{Primary and Corresponding author.} \\
Ryquo \\
\texttt{tianyi@ryquo.com}
\And
Nathan Huang \\
App-In Club \\
\texttt{nathan.huang@appinclub.org}
\And
Justin Tang \\
App-In Club \\
\texttt{justin@appinclub.org}
\AND
Wenqian Chen \\
App-In Club \\
\texttt{wenqian.chen@appinclub.org}
\And
Elsa Fan \\
Carnegie Mellon University \\
\texttt{elsaf@andrew.cmu.edu}
}

\begin{document}
\maketitle

\begin{abstract}
Large language models (LLMs) are now widely used as judges, yet their decisions can change under presentation choices that should be irrelevant. We study one such source of instability: candidate-order sensitivity in listwise factuality evaluation, where several answers can look similarly polished while differing substantially in hallucination risk. We introduce \method, an inference-time method that reruns the same factuality-first listwise prompt over multiple orderings of the same candidate set and aggregates the resulting scores, ranks, and uncertainty signals into a single consensus decision. On RewardBench~2 Factuality, the final seven-permutation aggregate ($K=7$) improves top-1 selection accuracy from 86.00\% to 91.33\% with GPT-5.4 and from 86.33\% to 89.67\% with Claude Sonnet~4.6. These results suggest that candidate order can be a meaningful source of factuality-judging error and that marginalizing over this nuisance variation can improve the reliability of LLM evaluation.
\end{abstract}

\section{Introduction}
LLM-as-a-judge has become a core evaluation primitive in modern NLP. Early systems such as G-Eval and PandaLM showed that large models can serve as practical reference-free evaluators and pairwise selectors \citep{liu2023geval,wang2023pandalm}, and strong proprietary models are now routinely used to rank candidate responses, approximate human preferences, audit downstream systems, and provide reward signals for post-training \citep{zheng2023judging,gu2024survey,li2024comprehensive}. At the same time, the reliability of these judges remains uncertain. A growing body of work documents position bias, rubric sensitivity, scale instability, and related forms of evaluation drift that can materially change judge decisions \citep{shi2025position,hong2026rulers,wang2025judgmentdist}. Recent work further argues that the relevant failure mode for judge-based selection is not global score correlation but within-prompt ranking quality: a judge can look good on aggregate metrics yet still make poor best-of-$N$ decisions \citep{landesberg2026bestofn,zhou2025jetts}.

This paper studies one source of instability: \emph{candidate-order sensitivity in listwise factuality evaluation}. RewardBench~2 explicitly supports rankings-based generative evaluation and includes a factuality subset designed to separate safer, better-calibrated responses from answers that sound confident but contain unsupported details \citep{malik2025rewardbench2}. This is exactly the regime in which ordering artifacts are dangerous. If a candidate is preferred only when it happens to appear first, the resulting judge is not measuring factuality robustly.

We ask a deliberately simple question: \emph{what if we treat candidate order as nuisance variation and average over it?} Our answer is \method, a judge that reruns a factuality-first listwise prompt over multiple permutations of the same candidate set and aggregates the results into a consensus score at runtime. The method uses no retrieval and no external verifier; it asks the same judge to evaluate the same candidates under shuffled orders and then extracts the stable signal.

Empirically, this intervention improves top-1 selection accuracy for both judge backbones on RewardBench~2 Factuality, with gains of 5.33 percentage points for GPT-5.4 and 3.33 percentage points for Claude Sonnet~4.6 in the final seven-permutation setting. Additional controls separate the effect of order perturbation from repeated canonical-order calls: averaging repeated evaluations in the canonical order gives little or no comparable benefit, while permutation consensus remains positive across reduced-$K$ settings. A pairwise transfer variant yields smaller but still positive gains in JudgeBench \citep{tan2024judgebench}. The contrast is informative: permutation consensus is especially effective in the \emph{multi-candidate factuality selection} setting it was designed for, but it is not designed to solve every judge setting.

Our contributions are as follows.
\begin{enumerate}[leftmargin=*,itemsep=2pt,topsep=2pt]
    \item We introduce a training-free permutation-consensus judge specialized to listwise factuality evaluation.
    \item We formalize the method as an order-robust consensus estimator and use a simple majority-vote analysis to motivate why consensus can reduce order noise, while explicitly noting that real permutation errors are correlated.
    \item We report RewardBench~2 Factuality improvements with two proprietary judge backbones, add reduced-$K$, repeated-canonical, and weight-sensitivity controls, and evaluate a smaller pairwise transfer variant on JudgeBench.
\end{enumerate}

\section{Related Work}

Broad surveys now frame LLM-as-a-judge research around four recurring design questions: how to construct judges, how to prompt or aggregate them, how to evaluate judges themselves, and how to control their biases in deployment \citep{gu2024survey,li2024comprehensive}. Our work is most aligned with the third and fourth categories: we treat judge reliability as an inference-time measurement problem rather than as a judge-training problem.

\paragraph{Building stronger evaluators.}
One common route to better LLM evaluation is to improve the evaluator itself, either through prompting, specialization, or model aggregation. Prompted systems such as G-Eval and PandaLM encode evaluation criteria or comparison procedures directly in the judge prompt \citep{liu2023geval,wang2023pandalm}, while MT-Bench and Chatbot Arena operationalize LLM judges for scalable conversational comparison \citep{zheng2023judging}. More recent work trains judge-specialized models, as in Prometheus~2 \citep{kim2024prometheus2}, or pools heterogeneous backbones, as in PoLL \citep{verga2024poll}. These approaches seek a stronger evaluator or a stronger panel of evaluators. Our work is complementary: we keep the backbone fixed and ask whether reliability can be improved by changing only the test-time protocol.

\paragraph{Judge quality depends on the downstream decision.}
A second line of work argues that judge quality should be evaluated with respect to the downstream decision the judge supports. RewardBench and RewardBench~2 study reward-model and judge accuracy on subtle preference, instruction-following, and factuality distinctions \citep{lambert2024rewardbench,malik2025rewardbench2}. JudgeBench shifts the focus further toward \emph{objective correctness}, constructing hard response pairs from reasoning-heavy source tasks such as MMLU-Pro, math, and coding \citep{tan2024judgebench,wang2024mmlupro}. JETTS evaluates judges in test-time-scaling settings such as reranking, beam search, and critique-based refinement, showing that judges can help some decision procedures more than others \citep{zhou2025jetts}. Closest aligned with our motivation, \citet{landesberg2026bestofn} argues that global agreement metrics can substantially overstate a judge's value for best-of-$N$ selection because they blur together prompt-level effects and within-prompt ranking quality. This perspective is central to our setting: listwise factuality evaluation is fundamentally a \emph{within-prompt selection} problem.

\paragraph{Bias, instability, and judge inference.}
A third line of work studies judges as noisy measurement instruments whose outputs can change under presentation choices that should be irrelevant. Early analyses of MT-Bench and Chatbot Arena already noted position and verbosity effects \citep{zheng2023judging}. Shi et al.\ provide a systematic study of position bias in both pairwise and listwise judging, showing that simple order changes can materially alter outcomes even for strong backbones \citep{shi2025position}. RULERS pushes this critique further by arguing that trustworthy judging requires locked rubrics, evidence-anchored scoring, and calibrated scales rather than prompt phrasing alone \citep{hong2026rulers}. Other work seeks to stabilize judge decisions without retraining the judge: \citet{wang2025judgmentdist} improve inference by using the full judgment distribution rather than greedy text outputs, while \citet{verga2024poll} reduce idiosyncratic model bias by pooling diverse judges. Our method aggregates over a different source of variation: not model diversity and not output-token uncertainty, but the order in which the same candidate set is presented.

\paragraph{What is still missing.}
Prior work has established that presentation order can bias LLM judges, including in listwise settings. Less is known about how to turn that diagnosis into a simple robustness procedure for \emph{top-1 factuality selection}, where several plausible candidates compete simultaneously and the downstream decision is the single answer to trust. This gap is particularly relevant when candidates differ less in fluency than in hallucination risk.

\section{Method}
\subsection{Problem setting}
Let $x$ be a prompt and $Y=\{y_1,\dots,y_n\}$ a set of candidate responses. The factuality target is order-invariant, but a deployed listwise judge consumes an ordered presentation of $Y$ and returns per-candidate scores and a winner. In practice, those outputs can depend on the presentation order. Our goal is to reduce this order sensitivity without changing the judge model or training a new evaluator.

\subsection{A skeptical factuality-first listwise prompt}
The direct baseline and \method share the same core prompt. The judge is instructed to rank candidates by factual reliability rather than by generic helpfulness, with particular emphasis on avoiding major factual error and unsupported specificity. The prompt asks for \emph{five} outputs for each candidate: a numeric score in $[0,100]$, a short rationale, a binary flag for major factual error, a binary flag for hallucinated specificity, and a binary flag for calibrated uncertainty. Calibrated uncertainty is treated as a weak positive signal only when it reflects appropriate caution rather than evasiveness. The two negative flags make factual-error and unsupported-specificity considerations explicit during each per-permutation scoring decision. We do not apply them a second time as external penalties in the final aggregation.

\subsection{Permutation-consensus aggregation}
\method runs the same prompt over $K$ orderings of the candidate list. Let $\pi^{(1)},\dots,\pi^{(K)}$ be candidate permutations. For each run $r$, the judge returns:
\begin{itemize}[leftmargin=*,itemsep=2pt,topsep=2pt]
    \item a score $s_i^{(r)} \in [0,100]$ for each candidate $i$,
    \item a full ranking, from which we derive a Borda-style contribution,
    \item a top-set indicator for the highest-scored candidate(s), and
    \item binary indicators for calibrated uncertainty, major error, and hallucinated specificity.
\end{itemize}

We map every response back to its original candidate identifier and aggregate the runs into four summary statistics:
\begin{align}
\bar{s}_i &= \frac{1}{K}\sum_{r=1}^K s_i^{(r)}, \\
B_i &= \frac{100}{K(n-1)}\sum_{r=1}^K \bigl(n-\operatorname{rank}_i^{(r)}\bigr), \\
v_i &= \frac{1}{K}\sum_{r=1}^K \frac{\mathbf{1}[i \in T^{(r)}]}{|T^{(r)}|}, \\
u_i &= \frac{1}{K}\sum_{r=1}^K \mathbf{1}[i \text{ marked calibrated uncertainty}],
\end{align}
where $T^{(r)}$ is the top-scoring set in run $r$ under the 0.5-point tie tolerance described below, and $\operatorname{rank}_i^{(r)}=1$ denotes the highest-ranked candidate. The final consensus score is
\begin{equation}
C_i = 0.50\bar{s}_i + 0.25 B_i + 0.20(100v_i) + 0.05(100u_i).
\label{eq:consensus}
\end{equation}
The winner is the candidate with maximal $C_i$; ties are retained when top scores fall within the fixed 0.5-point tolerance used in all experiments. Because $\bar{s}_i$, $B_i$, $100v_i$, and $100u_i$ all lie in $[0,100]$, the consensus score $C_i$ is itself a weighted average on the same scale, and the weights in Eq.~\ref{eq:consensus} sum to one. In the final RewardBench~2 runs we use $K=7$.

Equation~\ref{eq:consensus} matches the implementation used in the experiments reported below. The weights are a fixed heuristic rather than learned parameters; Section~\ref{sec:weight-sensitivity} ablates several reasonable alternatives. The weighting places most mass on two signals: per-permutation factuality score and order-robust relative rank. The uncertainty term is small by design: caution should help when it avoids fabricated detail, not dominate the decision. We do \emph{not} separately reweight the major-error and hallucinated-specificity flags in the final aggregation. In development, using those flags twice---once inside the judge's per-run scoring decision and again as an external penalty---tended to over-penalize cautious but incomplete responses.

\begin{algorithm}[h]
\caption{\method for one prompt $x$ with candidates $Y$}
\label{alg:pcfjudge}
\begin{algorithmic}[1]
\Require Prompt $x$, candidates $Y=\{y_1,\dots,y_n\}$, number of permutations $K$
\For{$r=1$ to $K$}
    \State sample or retrieve candidate permutation $\pi^{(r)}$
    \State run the factuality-first listwise judge on $(x,\pi^{(r)}(Y))$
    \State remap scores, ranks, and binary flags back to original candidate IDs
\EndFor
\For{each candidate $i$}
    \State compute $\bar{s}_i$, $B_i$, $v_i$, and $u_i$
    \State compute consensus score $C_i$ using Eq.~\ref{eq:consensus}
\EndFor
\State \Return candidate(s) with maximal $C_i$
\end{algorithmic}
\end{algorithm}

\subsection{Why consensus should help}
Our exact scoring rule is heuristic, but the core intuition admits a simple analysis. Suppose that under a random candidate ordering, the judge places the true best candidate first with probability $q>\tfrac{1}{2}$ and that these top-choice events are conditionally independent across permutations. Then majority vote over the top-choice identities already reduces error exponentially in $K$.

\begin{proposition}
Let $Z_r\in\{0,1\}$ indicate whether permutation run $r$ places the true best candidate first, with $\Pr(Z_r=1)=q>\tfrac{1}{2}$ and $Z_1,\dots,Z_K$ independent. If $K$ is odd and we choose the final winner by majority vote over the top choices, then
\[
\Pr\!\left(\sum_{r=1}^K Z_r \le \frac{K}{2}\right)
\le \exp\!\left(-2K\left(q-\frac{1}{2}\right)^2\right).
\]
\end{proposition}

This follows from Hoeffding's inequality \citep{hoeffding1963probability}. \method is richer than pure majority vote because it also aggregates within-run scores, full ranks, and calibrated uncertainty. The proposition should therefore be read as intuition rather than as a proof of Eq.~\ref{eq:consensus}.

Exact independence is unlikely in practice. All permutation runs share the same backbone, prompt, item, and candidate set, so their errors can be correlated. If the errors are perfectly correlated, repeated evaluation does not reduce error. We therefore use the reduced-$K$ and repeated canonical controls in Section~\ref{sec:controls} as empirical checks of whether order perturbations add signal beyond repeated calls.

\subsection{Pairwise transfer variant for JudgeBench}
JudgeBench is pairwise rather than listwise, so we use a separate transfer variant, \pairmethod. We use a separate name because this protocol is not the listwise algorithm in Algorithm~\ref{alg:pcfjudge}; it is an order-consensus adaptation for two-response evaluation. For a response pair $(y_A,y_B)$, the baseline direct judge scores the original order once. \pairmethod adds two safeguards. First, it evaluates both candidate orders and uses order-consensus as a disagreement signal. Second, it only accepts an order-based override when a separate keyed-judgment pass---which first resolves the underlying question and then compares the two responses against that resolved answer---confirms the same winner. To avoid a failure mode observed in development, the final variant skips keyed overrides on estimation-style prompts. We include JudgeBench to test scope, not to claim that this pairwise transfer variant is as strong as the main listwise method.

\section{Experimental Setup}
\subsection{Models}
We evaluate two proprietary judge backbones: GPT-5.4 via the OpenAI API \citep{openai2026gpt54} and Claude Sonnet~4.6 via the Anthropic API \citep{anthropic2026sonnet46}. We use the same backbone for both the direct baseline and the corresponding consensus method so that gains reflect the inference procedure rather than the base judge.

\subsection{RewardBench~2 Factuality}
Our main evaluation uses the public RewardBench~2 test split, specifically the Factuality subset \citep{malik2025rewardbench2}. Each item contains four candidate responses, making it a natural fit for listwise selection. For each backbone, we evaluate a fixed 300-example slice, with the direct baseline and all \method variants using the same slice for matched paired comparisons.

The direct baseline uses the same factuality-first listwise prompt as \method but evaluates only the canonical candidate order ($K=1$). \method uses predetermined permutations that are reused across items for reproducibility, with $K=7$ as the main setting. To reduce run-level noise in the final $K=7$ estimate, we execute the \method evaluation three times on the same slice and prompt configuration. For each item, we average the resulting per-candidate consensus scores across these executions and then select one final winner before computing the paired statistics in Table~\ref{tab:main-results}. The reported RewardBench~2 numbers are micro-averaged top-1 accuracies over the evaluated slice, expressed as percentages in the results tables. Tie rates were negligible in the final runs and in the diagnostic controls (mean top-tie size $\approx 1.00$). Diagnostic controls in Tables~\ref{tab:k-controls}--\ref{tab:weight-ablation} are reported as exact top-1 accuracy percentages so that each value has the same interpretation as the main RewardBench~2 accuracy columns.

\subsection{Mechanism and cost controls}
\label{sec:setup-controls}
We add three controls on the same RewardBench~2 slices. First, to separate order perturbation from repeated calls and aggregation, we run a repeated-canonical baseline: the judge sees the same canonical order $K$ times and we aggregate the repeated outputs with the same consensus rule. Second, to characterize cost, we evaluate \method at $K\in\{3,5,7\}$. Third, to test sensitivity to the heuristic weights in Eq.~\ref{eq:consensus}, we recompute winners under several fixed weight settings, including uniform weighting, score-only, rank-only, top-vote-only, and variants that remove the uncertainty term. These controls are computed from one fully logged diagnostic run for each backbone, so the reduced-$K$ and weight comparisons are made on matched raw judge outputs.

\subsection{JudgeBench transfer study}
JudgeBench contains objective response pairs derived from difficult source tasks, with public \texttt{gpt} and \texttt{claude} splits containing 350 and 270 unique pairs respectively \citep{tan2024judgebench}. Each JSON item includes the source bucket (e.g., \texttt{mmlu-pro-history}), the question, two candidate responses, and an objective label such as $A\!>\!B$. JudgeBench reports source-aware performance rather than only a single pooled accuracy, which is important because source tasks differ sharply in difficulty.

We evaluate fixed 100-pair slices from the public JudgeBench splits. Here, $N$ denotes the number of unique pairs. Following JudgeBench's source-aware reporting structure, we first compute accuracy within each source bucket represented in the slice and then macro-average those per-source values. This is why the reported percentage is not constrained to being a multiple of $1/N$.

\subsection{Metrics and significance}
For RewardBench~2 we report top-1 selection accuracy, expressed as a percentage, and paired improvement/regression counts relative to the direct baseline. For the main $K=7$ results we also report unchanged counts, since most examples agree with the direct judge. For the diagnostic control tables, all entries are exact top-1 accuracy percentages from a matched single run, while Table~\ref{tab:main-results} reports the final three-run aggregate described above. For JudgeBench we report the macro-averaged accuracy percentage described above. For the main RewardBench~2 slices we compute exact paired sign tests over the discordant examples (improved vs.~regressed) as a significance check.

\section{Results}

\subsection{Main RewardBench~2 results}
Table~\ref{tab:main-results} reports the main RewardBench~2 result. On RewardBench~2 Factuality, the final aggregate estimate for the $K=7$ configuration improves both judge backbones: GPT-5.4 improves from 86.00\% to 91.33\%, and Claude Sonnet~4.6 improves from 86.33\% to 89.67\%. The micro-average over the two evaluated slices improves from 86.17\% to 90.50\%.

\begin{table*}[t]
\centering
\small
\setlength{\tabcolsep}{9pt}
\begin{tabular}{lrrrrrr}
\toprule
Model & $N$ & Direct (\%) & \method (\%) & $\Delta$ (pp) & Imp./Reg. & Same \\
\midrule
GPT-5.4 & 300 & 86.00 & 91.33 & +5.33 & 21 / 5 & 274 \\
Claude 4.6 & 300 & 86.33 & 89.67 & +3.33 & 17 / 7 & 276 \\
\midrule
Micro avg. & 600 & 86.17 & 90.50 & +4.33 & 38 / 12 & 550 \\
\bottomrule
\end{tabular}
\caption{Main RewardBench~2 Factuality results. Accuracy columns are top-1 selection accuracy percentages; $\Delta$ is measured in percentage points. Direct uses one canonical candidate order. Each \method execution uses seven candidate-order permutations with the consensus rule in Eq.~\ref{eq:consensus}. The final $K=7$ \method numbers aggregate three separately executed $K=7$ evaluations on the same fixed 300-example slice. For each item, we average per-candidate consensus scores across the three executions before selecting one final winner; paired counts are then computed against the corresponding direct baseline.}
\label{tab:main-results}
\end{table*}

\begin{figure}[h]
    \centering
    \includegraphics[width=\columnwidth]{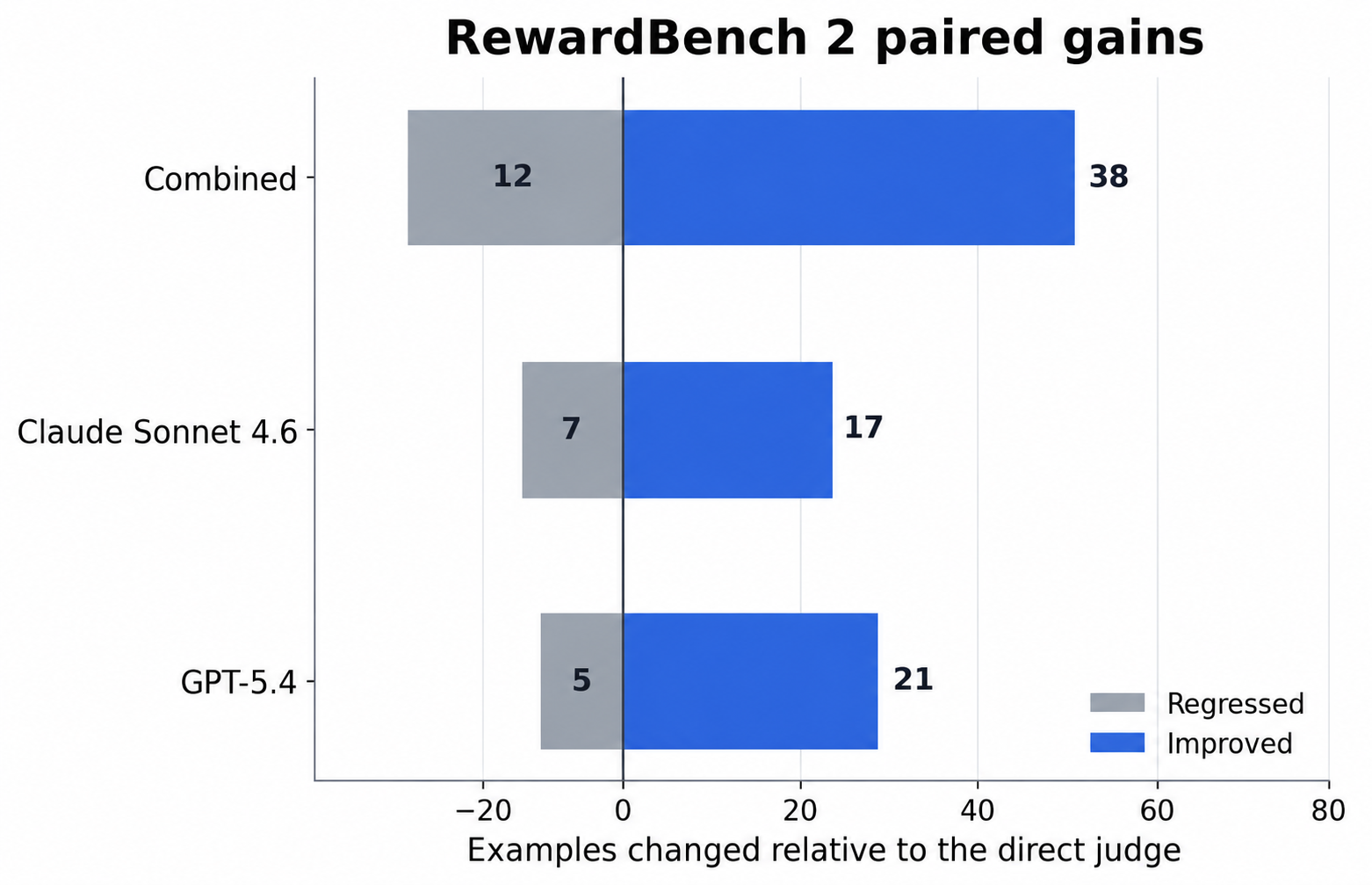}
    \caption{Paired comparison against the direct judge on the final 300-example RewardBench~2 Factuality slices. Bars show only changed examples; unchanged counts are reported in Table~\ref{tab:main-results}.}
    \label{fig:paired-gains}
\end{figure}

The paired counts show that the gains are not only changes in aggregate accuracy. GPT-5.4 improves on 21 examples and regresses on 5, giving an exact two-sided sign-test value of $p=0.0025$. Claude Sonnet~4.6 improves on 17 examples and regresses on 7, a positive but weaker paired signal ($p=0.064$). As a descriptive pooled check over the 600 evaluated item-backbone pairs, the discordant counts are 38 improvements versus 12 regressions ($p=3.1\times10^{-4}$). Figure~\ref{fig:paired-gains} visualizes the same paired asymmetry, while Table~\ref{tab:main-results} also shows that most examples remain unchanged.

The backbone-specific pattern is intentionally interpreted conservatively. Both models improve, but the paired evidence is stronger for GPT-5.4 than for Claude Sonnet~4.6. Together, these results support the narrower claim that order perturbation can recover some errors made by strong single-pass factuality judges in this setting.

\subsection{Cost and mechanism controls}
\label{sec:controls}
Table~\ref{tab:k-controls} reports the diagnostic control suite used to separate permutation consensus from repeated calls and to characterize the effect of $K$. The table uses one fully logged run per backbone so that $K=3$, $K=5$, $K=7$, and repeated-canonical baselines are directly comparable on the same raw judge outputs. Because this table is a matched single-run diagnostic, its $K=7$ entries are not expected to exactly match the three-run aggregate in Table~\ref{tab:main-results}.

\begin{table*}[t]
\centering
\small
\setlength{\tabcolsep}{7pt}
\begin{tabular}{llccc}
\toprule
Model & Protocol & $K=3$ (\%) & $K=5$ (\%) & $K=7$ (\%) \\
\midrule
\multirow{2}{*}{GPT-5.4}
& repeated canonical & 86.67 (+0.67) & 85.67 ($-$0.33) & 86.00 (+0.00) \\
& \method & 89.33 (+3.33) & 88.67 (+2.67) & 89.33 (+3.33) \\
\midrule
\multirow{2}{*}{Claude 4.6}
& repeated canonical & 86.00 ($-$0.33) & 87.00 (+0.67) & 86.67 (+0.33) \\
& \method & 87.00 (+0.67) & 87.67 (+1.33) & 88.00 (+1.67) \\
\bottomrule
\end{tabular}
\caption{Diagnostic cost and mechanism controls on RewardBench~2 Factuality. Entries are exact top-1 accuracy percentages, with absolute percentage-point change over the direct baseline in parentheses. Repeated canonical uses $K$ calls to the same candidate order; \method uses $K$ candidate-order permutations. This table is a matched single-run diagnostic control, while Table~\ref{tab:main-results} reports the final three-run aggregate for the main $K=7$ setting.}
\label{tab:k-controls}
\end{table*}

The repeated-canonical rows do not show a stable gain: additional calls to the same order do not reliably improve either backbone. In contrast, the permutation rows are positive across all reported $K$ values. GPT-5.4 already recovers most of the diagnostic-run gain at $K=3$, whereas Claude Sonnet~4.6 improves more gradually from $K=3$ to $K=7$. This suggests a practical tradeoff: $K=7$ is the base configuration used for the Table~\ref{tab:main-results} aggregate, but lower-$K$ settings may be preferable when inference cost is the binding constraint.

\subsection{Sensitivity to consensus weights}
\label{sec:weight-sensitivity}
Table~\ref{tab:weight-ablation} ablates the heuristic weights in Eq.~\ref{eq:consensus}. The proposed weights are not uniquely optimal, but the main conclusion is stable: a range of score-, rank-, and top-vote-based variants remain above the direct baseline on both backbones. This indicates that the improvement is not a fragile artifact of one hand-chosen coefficient vector.

\begin{table*}[t]
\centering
\small
\setlength{\tabcolsep}{5pt}
\begin{tabular}{lrrrrrr}
\toprule
Variant & $w_s$ & $w_B$ & $w_v$ & $w_u$ & GPT-5.4 (\%) & Claude 4.6 (\%) \\
\midrule
Proposed & 0.50 & 0.25 & 0.20 & 0.05 & 89.33 & 88.00 \\
Uniform & 0.25 & 0.25 & 0.25 & 0.25 & 90.00 & 87.00 \\
Score only & 1.00 & 0.00 & 0.00 & 0.00 & 88.67 & 87.00 \\
Rank only & 0.00 & 1.00 & 0.00 & 0.00 & 89.00 & 88.00 \\
Top vote only & 0.00 & 0.00 & 1.00 & 0.00 & 89.00 & 88.33 \\
No uncertainty & 0.50 & 0.27 & 0.23 & 0.00 & 89.33 & 88.33 \\
Score/rank & 0.50 & 0.50 & 0.00 & 0.00 & 89.33 & 88.00 \\
Score/top & 0.60 & 0.00 & 0.40 & 0.00 & 89.33 & 88.67 \\
\bottomrule
\end{tabular}
\caption{Weight sensitivity on the same diagnostic runs as Table~\ref{tab:k-controls}. Entries are exact top-1 accuracy percentages. Columns show the weights on mean score ($w_s$), Borda rank ($w_B$), top-set frequency ($w_v$), and calibrated uncertainty ($w_u$). Direct baselines are 86.00\% for GPT-5.4 and 86.33\% for Claude Sonnet~4.6.}
\label{tab:weight-ablation}
\end{table*}

We keep Eq.~\ref{eq:consensus} as the main rule because it is the final configuration used for the main experiments and balances score, rank, top-set agreement, and a small uncertainty signal. The ablation suggests, however, that the exact 0.50/0.25/0.20/0.05 split should not be over-interpreted. In these runs, the robust signal comes primarily from aggregating over order perturbations, not from tuning a delicate set of weights.

\subsection{JudgeBench transfer}
Table~\ref{tab:judgebench} reports the transfer study. The gains are smaller than on RewardBench~2, but they remain positive for both backbones: +3.24 percentage points for Claude Sonnet~4.6 and +2.70 percentage points for GPT-5.4. This difference is consistent with the narrower scope of the method. JudgeBench is an objective, pairwise, domain-diverse benchmark rather than the listwise factuality setting \method was designed for. The transfer experiment therefore serves as a boundary-condition check: order-robust judging still helps, but the strongest evidence remains the listwise factuality setting where candidate-order instability is directly tied to the downstream decision.

\begin{table}[t]
\centering
\small
\setlength{\tabcolsep}{3pt}
\begin{tabular*}{\columnwidth}{@{\extracolsep{\fill}}lrrrr@{}}
\toprule
Model & $N$ & Direct (\%) & \pairmethod (\%) & $\Delta$ (pp) \\
\midrule
Claude 4.6 & 100 & 79.09 & 82.33 & +3.24 \\
GPT-5.4 & 100 & 76.21 & 78.91 & +2.70 \\
\bottomrule
\end{tabular*}
\caption{JudgeBench transfer results. $N$ counts unique response pairs, while the reported value is the macro-averaged accuracy percentage over the source buckets present in the 100-pair slice, so the percentages need not be multiples of $1/N$. \pairmethod uses our order-swapped pairwise protocol with keyed confirmation before accepting an override.}
\label{tab:judgebench}
\end{table}

\subsection{Development ablations and lessons}
\begin{figure}[h]
    \centering
    \includegraphics[width=\columnwidth]{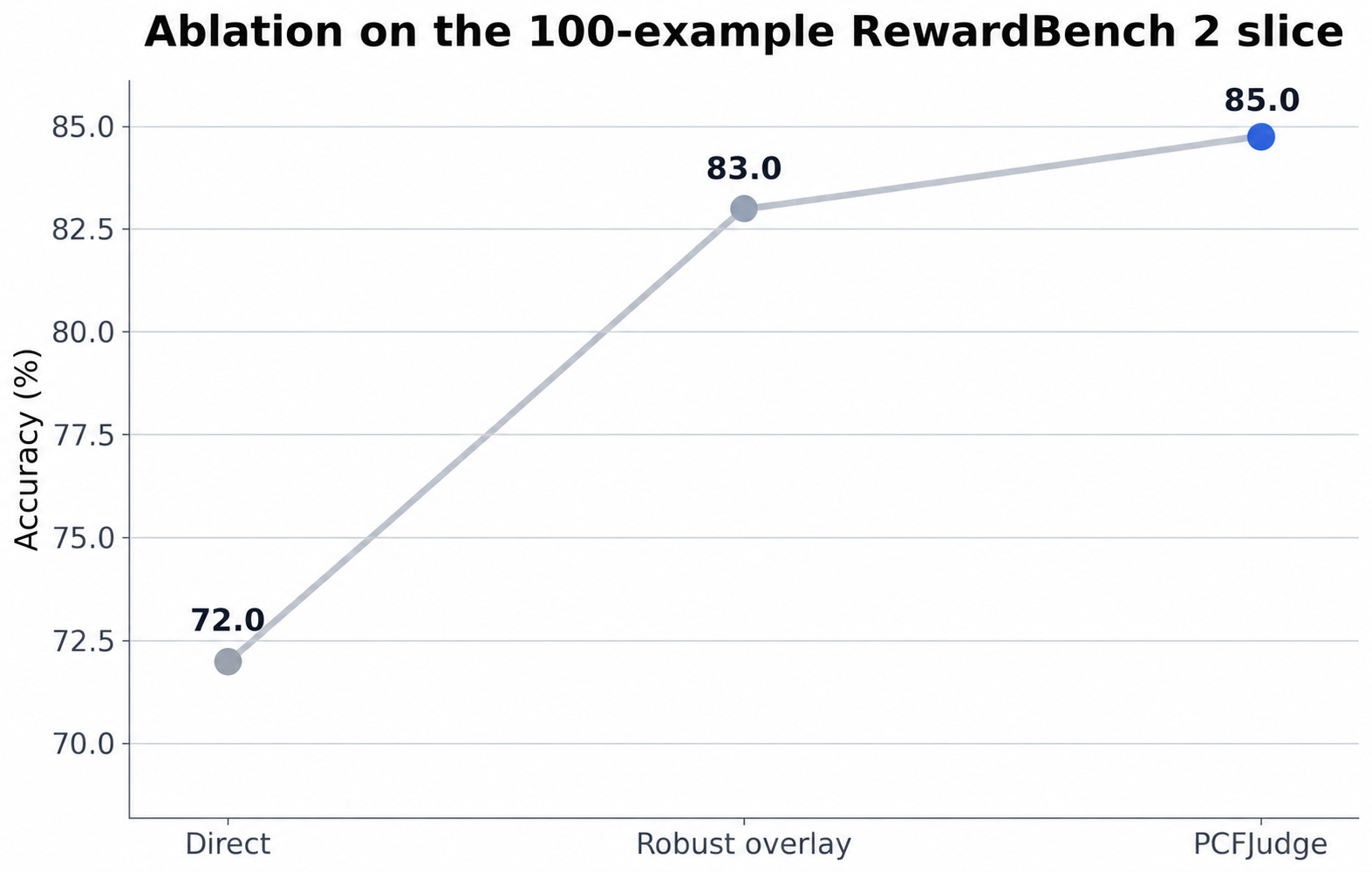}
    \caption{Development ablation on a fixed 100-example GPT-5.4 RewardBench~2 Factuality slice. Values are exact top-1 accuracy percentages on the development slice. Most of the recoverable error is removed by permutation consensus itself; extra overlay logic helps less than simply trusting the consensus ranker.}
    \label{fig:ablation}
\end{figure}

Development ablations informed the final design. Figure~\ref{fig:ablation} reports an earlier comparable ablation on a fixed 100-example GPT-5.4 RewardBench~2 Factuality development slice. The ``robust overlay'' variant was a heavier two-stage design that first produced a permutation-consensus ranking and then added an additional arbitration pass over the most plausible winners. It recovered much of the gain over direct judging, but the simpler permutation-consensus ranker still performed better. This is why the final method keeps the core decision rule simple rather than stacking additional arbitration layers on top of it.

Earlier 50-example development slices showed the same pattern more noisily: anchor ladders, panel arbitration, and evidence-backed overrides did not reliably improve over simpler consensus ranking, and some variants regressed. These attempts suggested that additional judge stages do not automatically add independent signal. In our setting, directly targeting order variation appeared more effective than introducing extra arbiters.

Together, Figure~\ref{fig:ablation} and the controls in Tables~\ref{tab:k-controls}--\ref{tab:weight-ablation} support a modest lesson: a targeted intervention on a concrete nuisance variable can be more useful than increasingly elaborate meta-judging pipelines.

\subsection{Qualitative patterns}
Table~\ref{tab:qualitative} summarizes representative qualitative patterns from prediction-file inspection. These patterns are diagnostic rather than a complete error taxonomy, but they help explain the kinds of cases where the method can help. The central effect is not extra world knowledge. Rather, \method is less willing to reward an answer whose advantage depends on a particular presentation order.

\begin{table*}[t]
\centering
\small
\renewcommand{\arraystretch}{1.8}
\setlength{\tabcolsep}{6pt}
\begin{tabularx}{\textwidth}{
>{\raggedright\arraybackslash}p{2.1cm}
>{\raggedright\arraybackslash}X
>{\raggedright\arraybackslash}X
>{\raggedright\arraybackslash}X
}
\toprule
\textbf{Pattern} & \textbf{Observed direct-judge behavior} & \textbf{Observed PCFJudge recovery} & \textbf{Why it matters} \\
\midrule
Unsupported specificity
& Polished answers that add precise settings, dates, or source claims without solid support
& Safer answers whose advantage survives across permutations
& Illustrates how consensus can reduce reward for hallucinated detail rather than merely smoothing scores \\
\midrule
Calibrated narrow correction
& Broader but speculative responses that sound more complete
& Narrower responses that correctly state that evidence is limited or absent
& Aligns the judge with factual reliability rather than informativeness style \\
\midrule
Near-tie candidate sets
& Small stylistic differences can dominate when all candidates share the same broad factual stance
& Limited changes unless an order-stable preference emerges
& Explains why improvements are concentrated on genuinely unstable cases \\
\midrule
Transfer failure mode
& Pairwise estimation or ballpark prompts where an auxiliary checker can overcommit
& Final transfer variant suppresses these overrides
& Clarifies why JudgeBench gains are positive but smaller than RewardBench~2 \\
\bottomrule
\end{tabularx}
\caption{Representative qualitative patterns from prediction-file inspection. The main benefit of PCFJudge is not extra knowledge; it is greater reluctance to reward answers whose advantage depends on a particular candidate order.}
\label{tab:qualitative}
\end{table*}

The first two rows of Table~\ref{tab:qualitative} capture observed RewardBench~2 success modes in corrected cases. Direct judging can overweight polished answers that add unsupported specifics on product-setting, website-grounded, or niche-source prompts. Permutation consensus can favor the safer and better-calibrated response when that preference remains stable across orderings. In these cases, the method is not making the judge more encyclopedic; it is making the judge less vulnerable to rhetorically attractive but unstable winners. The last two rows help explain why transfer is positive but smaller on JudgeBench. RewardBench~2 Factuality often presents several plausible answers whose key difference is unsupported specificity, whereas many JudgeBench pairs turn on objective task solving. This contrast is consistent with the intended scope of the method.

\section{Conclusion}
\method suggests that arbitrary candidate order is a consequential nuisance variable in listwise factuality selection. By marginalizing over order instead of trusting a single presentation, the same judge backbone can make more stable choices without finetuning, retrieval, or a separate verifier. The repeated-canonical control indicates that the benefit is not simply the result of making more calls; the order perturbations themselves provide useful signal. The method still has real costs and does not remove the need for broader audits, but the results suggest that order-robustness is a useful design consideration for future factuality-judging pipelines.

\section*{Limitations}
The main evidence comes from fixed API-budgeted slices rather than full benchmark sweeps, so larger runs would better characterize variance across slices and domains. \method also increases inference cost by a factor of $K$ in its basic form; the reduced-$K$ controls suggest that $K=3$ or $K=5$ can be useful when cost is binding, but the extra calls remain a practical deployment cost. The consensus weights in Eq.~\ref{eq:consensus} are heuristic: Table~\ref{tab:weight-ablation} suggests that the method is not brittle to the chosen weights, but does not establish that they are optimal. Proposition~1 assumes independent permutation errors, whereas repeated calls to the same model, prompt, and candidate set are likely correlated; such correlation can limit consensus gains. Finally, the strongest evidence is on RewardBench~2 Factuality, while JudgeBench shows smaller pairwise transfer. Our study isolates presentation-order variation and does not address issues such as benchmark validity, label noise, or hidden contamination.

\section*{Broader Impact and Ethical Considerations}
This work focuses on evaluation rather than direct user-facing generation, but evaluation still shapes what behaviors are rewarded during model development and deployment. More stable factuality judges can reduce incentives to reward confident fabrication, make best-of-$N$ pipelines less sensitive to arbitrary prompt order, and make judge failures easier to inspect when models are iterated rapidly. At the same time, a factuality-first judge may over-prefer terse caution, under-credit partially correct exploratory answers, or entrench a benchmark's notion of acceptable uncertainty if it is deployed outside its intended scope.

Because \method averages multiple judge calls, it also increases API usage and, if adopted uncritically, may further concentrate the evaluative power in proprietary systems. We therefore view it as a limited-scope reliability layer that should be paired with domain-specific audits, human oversight, and benchmark validation rather than treated as a stand-alone arbiter of truth. If evaluation infrastructure shapes what future models are rewarded to learn, then making that infrastructure less arbitrary is a practical step toward reducing one pathway by which hallucinations are reinforced.

\bibliography{custom}

@misc{anthropic2026sonnet46,
      author = {{Anthropic}},
      title = {Introducing Claude Sonnet 4.6},
      year = {2026},
      howpublished = {\url{https://www.anthropic.com/news/claude-sonnet-4-6}},
}

@misc{gu2024survey,
      title={A Survey on LLM-as-a-Judge}, 
      author={Jiawei Gu and Xuhui Jiang and Zhichao Shi and Hexiang Tan and Xuehao Zhai and Chengjin Xu and Wei Li and Yinghan Shen and Shengjie Ma and Honghao Liu and Saizhuo Wang and Kun Zhang and Yuanzhuo Wang and Wen Gao and Lionel Ni and Jian Guo},
      year={2025},
      eprint={2411.15594},
      archivePrefix={arXiv},
      primaryClass={cs.CL},
      url={https://arxiv.org/abs/2411.15594}, 
}

@article{hoeffding1963probability,
     ISSN = {01621459, 1537274X},
     URL = {http://www.jstor.org/stable/2282952},
     author = {Wassily Hoeffding},
     journal = {Journal of the American Statistical Association},
     number = {301},
     pages = {13--30},
     publisher = {[American Statistical Association, Taylor & Francis, Ltd.]},
     title = {Probability Inequalities for Sums of Bounded Random Variables},
     urldate = {2026-03-20},
     volume = {58},
     year = {1963}
}

@misc{shi2025position,
      title={Judging the Judges: A Systematic Study of Position Bias in LLM-as-a-Judge}, 
      author={Lin Shi and Chiyu Ma and Wenhua Liang and Xingjian Diao and Weicheng Ma and Soroush Vosoughi},
      year={2025},
      eprint={2406.07791},
      archivePrefix={arXiv},
      primaryClass={cs.CL},
      url={https://arxiv.org/abs/2406.07791}, 
}

@misc{hong2026rulers,
      title={RULERS: Locked Rubrics and Evidence-Anchored Scoring for Robust LLM Evaluation}, 
      author={Yihan Hong and Huaiyuan Yao and Bolin Shen and Wanpeng Xu and Hua Wei and Yushun Dong},
      year={2026},
      eprint={2601.08654},
      archivePrefix={arXiv},
      primaryClass={cs.CL},
      url={https://arxiv.org/abs/2601.08654}, 
}

@misc{kim2024prometheus2,
      title={Prometheus 2: An Open Source Language Model Specialized in Evaluating Other Language Models}, 
      author={Seungone Kim and Juyoung Suk and Shayne Longpre and Bill Yuchen Lin and Jamin Shin and Sean Welleck and Graham Neubig and Moontae Lee and Kyungjae Lee and Minjoon Seo},
      year={2024},
      eprint={2405.01535},
      archivePrefix={arXiv},
      primaryClass={cs.CL},
      url={https://arxiv.org/abs/2405.01535}, 
}

@misc{landesberg2026bestofn,
      title={When LLM Judge Scores Look Good but Best-of-N Decisions Fail}, 
      author={Eddie Landesberg},
      year={2026},
      eprint={2603.12520},
      archivePrefix={arXiv},
      primaryClass={cs.LG},
      url={https://arxiv.org/abs/2603.12520}, 
}

@misc{malik2025rewardbench2,
      title={RewardBench 2: Advancing Reward Model Evaluation}, 
      author={Saumya Malik and Valentina Pyatkin and Sander Land and Jacob Morrison and Noah A. Smith and Hannaneh Hajishirzi and Nathan Lambert},
      year={2025},
      eprint={2506.01937},
      archivePrefix={arXiv},
      primaryClass={cs.CL},
      url={https://arxiv.org/abs/2506.01937}, 
}

@misc{openai2026gpt54,
      author = {{OpenAI}},
      title = {Introducing GPT‑5.4},
      year = {2026},
      howpublished = {\url{https://openai.com/index/introducing-gpt-5-4/}},
}

@misc{tan2024judgebench,
      title={JudgeBench: A Benchmark for Evaluating LLM-based Judges}, 
      author={Sijun Tan and Siyuan Zhuang and Kyle Montgomery and William Y. Tang and Alejandro Cuadron and Chenguang Wang and Raluca Ada Popa and Ion Stoica},
      year={2025},
      eprint={2410.12784},
      archivePrefix={arXiv},
      primaryClass={cs.AI},
      url={https://arxiv.org/abs/2410.12784}, 
}

@misc{verga2024poll,
      title={Replacing Judges with Juries: Evaluating LLM Generations with a Panel of Diverse Models}, 
      author={Pat Verga and Sebastian Hofstatter and Sophia Althammer and Yixuan Su and Aleksandra Piktus and Arkady Arkhangorodsky and Minjie Xu and Naomi White and Patrick Lewis},
      year={2024},
      eprint={2404.18796},
      archivePrefix={arXiv},
      primaryClass={cs.CL},
      url={https://arxiv.org/abs/2404.18796}, 
}

@misc{wang2024mmlupro,
      title={MMLU-Pro: A More Robust and Challenging Multi-Task Language Understanding Benchmark}, 
      author={Yubo Wang and Xueguang Ma and Ge Zhang and Yuansheng Ni and Abhranil Chandra and Shiguang Guo and Weiming Ren and Aaran Arulraj and Xuan He and Ziyan Jiang and Tianle Li and Max Ku and Kai Wang and Alex Zhuang and Rongqi Fan and Xiang Yue and Wenhu Chen},
      year={2024},
      eprint={2406.01574},
      archivePrefix={arXiv},
      primaryClass={cs.CL},
      url={https://arxiv.org/abs/2406.01574}, 
}

@misc{wang2025judgmentdist,
      title={Improving LLM-as-a-Judge Inference with the Judgment Distribution}, 
      author={Victor Wang and Michael J. Q. Zhang and Eunsol Choi},
      year={2025},
      eprint={2503.03064},
      archivePrefix={arXiv},
      primaryClass={cs.CL},
      url={https://arxiv.org/abs/2503.03064}, 
}

@misc{zhou2025jetts,
      title={Evaluating Judges as Evaluators: The JETTS Benchmark of LLM-as-Judges as Test-Time Scaling Evaluators}, 
      author={Yilun Zhou and Austin Xu and Peifeng Wang and Caiming Xiong and Shafiq Joty},
      year={2025},
      eprint={2504.15253},
      archivePrefix={arXiv},
      primaryClass={cs.CL},
      url={https://arxiv.org/abs/2504.15253}, 
}

@misc{liu2023geval,
      title={G-Eval: NLG Evaluation using GPT-4 with Better Human Alignment}, 
      author={Yang Liu and Dan Iter and Yichong Xu and Shuohang Wang and Ruochen Xu and Chenguang Zhu},
      year={2023},
      eprint={2303.16634},
      archivePrefix={arXiv},
      primaryClass={cs.CL},
      url={https://arxiv.org/abs/2303.16634}, 
}

@misc{wang2023pandalm,
      title={PandaLM: An Automatic Evaluation Benchmark for LLM Instruction Tuning Optimization}, 
      author={Yidong Wang and Zhuohao Yu and Zhengran Zeng and Linyi Yang and Cunxiang Wang and Hao Chen and Chaoya Jiang and Rui Xie and Jindong Wang and Xing Xie and Wei Ye and Shikun Zhang and Yue Zhang},
      year={2024},
      eprint={2306.05087},
      archivePrefix={arXiv},
      primaryClass={cs.CL},
      url={https://arxiv.org/abs/2306.05087}, 
}

@misc{zheng2023judging,
      title={Judging LLM-as-a-Judge with MT-Bench and Chatbot Arena}, 
      author={Lianmin Zheng and Wei-Lin Chiang and Ying Sheng and Siyuan Zhuang and Zhanghao Wu and Yonghao Zhuang and Zi Lin and Zhuohan Li and Dacheng Li and Eric P. Xing and Hao Zhang and Joseph E. Gonzalez and Ion Stoica},
      year={2023},
      eprint={2306.05685},
      archivePrefix={arXiv},
      primaryClass={cs.CL},
      url={https://arxiv.org/abs/2306.05685}, 
}

@misc{li2024comprehensive,
      title={LLMs-as-Judges: A Comprehensive Survey on LLM-based Evaluation Methods}, 
      author={Haitao Li and Qian Dong and Junjie Chen and Huixue Su and Yujia Zhou and Qingyao Ai and Ziyi Ye and Yiqun Liu},
      year={2024},
      eprint={2412.05579},
      archivePrefix={arXiv},
      primaryClass={cs.CL},
      url={https://arxiv.org/abs/2412.05579}, 
}

@misc{lambert2024rewardbench,
      title={RewardBench: Evaluating Reward Models for Language Modeling}, 
      author={Nathan Lambert and Valentina Pyatkin and Jacob Morrison and LJ Miranda and Bill Yuchen Lin and Khyathi Chandu and Nouha Dziri and Sachin Kumar and Tom Zick and Yejin Choi and Noah A. Smith and Hannaneh Hajishirzi},
      year={2024},
      eprint={2403.13787},
      archivePrefix={arXiv},
      primaryClass={cs.LG},
      url={https://arxiv.org/abs/2403.13787}, 
}

\end{document}